\pdfoutput=1

\documentclass[11pt]{article}
\usepackage[hyphens]{url}
\usepackage[preprint]{joddash}
\usepackage{booktabs,graphicx,xspace,amsmath}

\usepackage{times}
\usepackage{latexsym}
\usepackage{newtxtext}

\usepackage[T1]{fontenc}
\usepackage{lmodern}

\usepackage[utf8]{inputenc}

\usepackage{microtype}

\usepackage{inconsolata}
\usepackage{enumitem}

\usepackage{graphicx}
\usepackage{listings}    
\usepackage{fvextra}  
\usepackage{subcaption}
\usepackage{tcolorbox} 
\tcbuselibrary{listings,skins,breakable}

\newcommand\model[1]{\textsc{#1}}

\newcommand{\unforcednodef}{\model{NoCtx}\xspace}
\newcommand{\forcednodef}{\model{Assumed}\xspace}
\newcommand{\oneshot}{\model{Assumed-IcE}\xspace}
\newcommand{\unforceddef}{\model{Def}\xspace}
\newcommand{\forceddef}{\model{Assumed-Def}\xspace}
\newcommand{\ice}{\model{Ice}\xspace}

\newcommand{\gpt}{\texttt{GPT-4o}\xspace}
\newcommand{\llama}{\texttt{Llama}\xspace}
\newcommand{\longllama}{\texttt{Llama-3.2-3B-Instruct}\xspace}

\newcommand\dataset[1]{{\color{violet}\textbf{#1}}\xspace}
\newcommand\type[1]{{\color{green!30!black}\textit{#1}}\xspace}
\newcommand\category[1]{{\color{green!30!black}\textbf{#1}}\xspace}

\newcommand\targeting{\category{Targeting}}
\newcommand\expression{\category{Expression}}

\newcommand\historical{\type{Historical}}
\newcommand\genocidal{\type{Genocidal}}
\newcommand\condoning{\type{Condoning Terrorism}}
\newcommand\bullying{\type{Bullying}}
\newcommand\physical{\type{Assault}}
\newcommand\discrimination{\type{Discrimination}}
\newcommand\destruction{\type{Destruction}}
\newcommand\suppression{\type{Suppression}}
\newcommand\denigration{\type{Denigration}}

\newcommand\harassment{\category{Harassment}}
\newcommand\assault{\category{Assault}}
\newcommand\vandalism{\category{Vandalism}}

\usepackage{capt-of}

\title{Evaluating Large Language Models for Antisemitic Incident Classification}

\author{
\textbf{Karina Halevy\textsuperscript{1}},
\textbf{Julia Mendelsohn\textsuperscript{2}},
\textbf{Chan Young Park\textsuperscript{3}},\\
\textbf{Yulia Tsvetkov\textsuperscript{4}},
\textbf{Maarten Sap\textsuperscript{1,5}}
\\
 \textsuperscript{1}Carnegie Mellon University,
\textsuperscript{2}University of Maryland,
\textsuperscript{3}Microsoft Research,\\
\textsuperscript{4}University of Washington,
\textsuperscript{5}Allen Institute for Artificial Intelligence\\
\small{
\textbf{Correspondence:} \href{mailto:khalevy@andrew.cmu.edu}{khalevy@andrew.cmu.edu}
}
}

\begin{document}

\maketitle
\vspace{-1em}
\begin{abstract}
Addressing hate and violence in society requires timely detection of hateful events from public reporting, but automated identification of hateful events remains underexplored. We introduce the task of \textit{hateful event detection} and investigate the ability of AI systems, specifically large language models (LLMs), to discover and classify reports of antisemitic events with fine-grained labels. We evaluate OpenAI's \gpt and Meta's \longllama on multiple expert-annotated datasets containing antisemitic event descriptions from news articles, civil society reports, and official records. We show that LLMs, particularly \gpt, have potential for this task, but substantial improvement is needed. Providing clear term definitions and in-context examples in prompts can improve performance: definitions are most helpful for rhetoric-oriented events (e.g. classical antisemitic tropes), while examples help label action-oriented events (e.g. physical assault). A case study of college newspapers demonstrates that LLMs can help surface relevant real-world events, supporting early monitoring and intervention. Overall, our findings highlight both opportunities and critical gaps in AI’s ability to recognize complex harms and underscore the need for collaborative efforts among AI developers, policymakers, and civil society to design models, implement robust evaluation, and develop policy frameworks for defining and combating hate efficiently and effectively.
\end{abstract}

\section{Introduction}
Hate and violence in society are not only individual tragedies but also indicators of broader social harm. Detecting hateful events---broadly defined as crimes, threats, or encouragement of crimes motivated by bias---in descriptions from news articles, civil society reports, and official records is crucial for monitoring societal trends and protecting targeted communities \citep{justice}. 
However, the growing volume of such reports makes comprehensive, timely monitoring increasingly difficult to carry out by human analysts alone, and there is a clear need for automated tools.

Prior computational approaches for analyzing real-world hate have largely focused on detecting hateful or toxic language, typically in social media posts. While valuable, these approaches are limited for jointly monitoring online and offline harm---they center on \textit{speech} rather than \textit{events}. This excludes many forms of bias-motivated harm such as physical violence, vandalism, and discrimination, which may contain no explicitly hateful language. 

This work addresses these limitations by introducing the novel task of \textit{hateful event detection}---characterizing both online and offline hate incidents as described in textual reports (e.g. distilling mentions of hateful events from streams of social media posts, press releases, news articles). This task is conceptually distinct from, yet complementary to, hate speech detection, which focuses on explicitly hateful language. This task moreover emphasizes \textit{fine-grained classification}, requiring computational models to distinguish among specific forms of harm rather than solely assigning overly broad labels---coarse labels obscure important distinctions among different types of targets and harm, limiting their utility for practitioners who must decide when, where, and how to intervene. This task facilitates a structured and actionable representation of real-world hate that better aligns with the goals of policymakers, educators, and civil society organizations. Stakeholders such as NGOs, journalists, law enforcement officials, and social media platform moderators would also benefit from a tool that accurately performs this task; a screening tool that automatically categorizes hateful events would also expand the scope of events detected and addressed by formal reporting systems alone.

We study hateful event detection with a case study on antisemitic incidents, as antisemitism is a particularly complex and socially consequential domain. Definitions of antisemitism are contested, with a high degree of subjectivity in the perception of events as antisemitic \citep{waxman2022arguing}. Furthermore, its manifestations span a wide range of behaviors, from coded rhetoric communicating historical tropes and conspiracy narratives, to harassment, vandalism of Jewish institutions, and physical violence. Such characteristics make the identification of antisemitic events particularly technically challenging for automated systems, as they must interpret intent, context, and history. Focusing on antisemitism thus offers insights both into current AI models' capabilities and the broader challenges of detecting complex forms of social harms. Moreover, it is especially urgent to investigate automated approaches to help address antisemitism given its high prevalence around the world, sharp increase in recent years, and deadly consequences.

This work offers multiple contributions to both AI research and hate detection studies. First, we introduce and formalize the task of hateful event detection, emphasizing the role of fine-grained taxonomies in capturing concrete forms of harm. Second, we assemble and release new resources for studying this task, including a dataset of reports about antisemitic incidents from the AMCHA initiative and a synthetic contrast set designed to test AI models' ability to distinguish antisemitic incidents from non-antisemitic but Jewish-related events. Third, we systematically evaluate state-of-the-art AI systems, specifically large language models (LLMs), on this task. Beyond examining overall performance, we also assess how varying the information provided in model prompts shape their ability to reason about historically and socially-grounded harms. 

Our experiments reveal that there is indeed potential for LLMs to aid humans in real-world hate monitoring. However, there is a substantial need for improvement before practitioners can fully rely on these models, as they are limited in their understanding of historical context, antisemitic symbolism, and the context-dependent ways in which antisemitism can be enacted in everyday interpersonal interactions. We find that including additional information in model prompts can meaningfully improve performance. In particular, precise taxonomy component definitions help LLMs classify rhetoric-oriented events, and in-context examples of manually-labeled reports helps LLMs classify action-oriented events. Finally, we demonstrate the practical relevance of hateful event detection by estimating antisemitic incident prevalence from recent university news articles. 
While preliminary due to the limited availability of human-annotated campus news articles, this case study shows that LLMs can help surface potentially relevant events for human review, streamlining otherwise costly data labeling efforts.

Taken together, our findings suggest that specificity in term definitions and annotation guidelines is central to building generalizable systems for detecting social harm. Beyond antisemitism, we argue that progress in computational approaches to hate and violence will require closer collaboration between AI researchers, domain experts, and civil society to develop shared taxonomies that are both technically usable and socially meaningful.

\begin{figure}[t]
    \centering
    \includegraphics[width=.5\columnwidth]{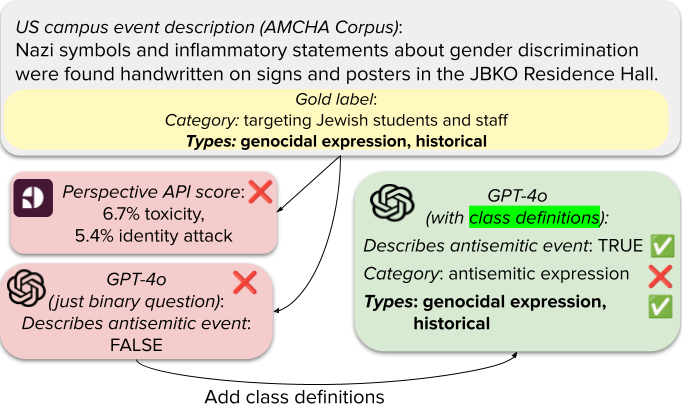}
    \caption{We introduce the novel task of fine-grained hateful event detection. This task is distinct from hate speech or toxicity detection: an example report describing an antisemitic incident receives low scores from Perspective API, a popular toxicity detection tool at the time of our dataset's collection, now being sunset. When only prompted with the event description, the LLM \gpt does not identify this example incident as antisemitic. Providing explicit definitions of fine-grained types of harm enables the model to correctly label the incident.}
    \label{fig:fig1}
\end{figure}

\section{Background}
\paragraph{Antisemitism}
Antisemitism refers to prejudice, discrimination, hostility, or hatred toward Jews \citep{ihra}.\footnote{\url{https://en.wikipedia.org/wiki/Antisemitism}} Antisemitism may be expressed through words or actions targeting Jewish individuals, communities, property, institutions, or religious sites \citep{ihra,jda}. It is an ancient phenomenon that predates modern notions of racism or religious discrimination. Because Jewish identity encompasses faith, ancestry, ethnicity, and peoplehood, antisemitism often blends religious, racial, nationalistic, and conspiratorial elements. Antisemitism has increased steadily worldwide since the early 2000s \citep{judaken2008so}, intensified throughout the 2010s and the COVID-19 pandemic \citep{lafreniere2023explaining}, and escalated sharply in the United States and globally since October 2023 \citep{nyt_antisemitism_urgent_problem_2025,time_sutherland_antisemitism_political_violence_2025}.
Survey data from 2024 reveals increasing antisemitic attitudes around the world \citep{adl_2024_world_attitudes}. In the United States, beliefs in anti-Jewish tropes are at the highest levels since 1964, particularly rising among younger generations \citep{adl_2024_america_attitudes}.

This change in antisemitic attitudes is accompanied by concerning trends in violence against Jews. In 2025 alone: a Jewish U.S. governor's residence was set ablaze during Passover \citep{politico_shapiro_arson_2025}, a shooting outside a Jewish museum in Washington D.C. during a Jewish community event killed two people \citep{nyt_jewish_museum}, a Jewish woman marching for the return of Israeli hostages was murdered in a firebombing attack in Boulder, Colorado \citep{nyt_boulder}, two people were killed on Yom Kippur in a synagogue attack in Manchester, U.K. \citep{reuters_antisemitic_attack_rise_2025}, and fifteen Jews were murdered at a Hanukkah celebration on Bondi Beach in Sydney, Australia \citep{abc_bondi}. Violent and hateful incidents against Jews have become commonplace. According to FBI hate crime statistics from 2024, 18\% of all U.S. hate crimes and 69\% of all religiously-motivated hate crimes have targeted Jews, despite Jewish people being just 2\% of the population \citep{fbi_hate_crime_data_2024}. Antisemitic hate crimes have increased 5.8\% just between 2023 and 2024 alone \citep{jewishfederations_fbi_data_2025}. Jewish community members regularly experience antisemitism and are concerned for their day-to-day safety \citep{adl2025_experience}.

\paragraph{Detecting Antisemitism}

The widespread nature of antisemitism and its severe societal consequences make it imperative to develop automated tools that can rapidly track both online expressions and offline events. While hate speech and toxic language detection have been popular tasks in natural language processing (NLP) research for over a decade, \citep{warner2012detecting,schmidt}, antisemitism detection is relatively understudied. Some NLP resources focused on harmful language detection include antisemitism as one of a broader set of hate ideologies \citep{sap2020socialbiasframes,elsherief2021latent,vargas2022hatebr,toxigen}.

Computational approaches for antisemitism detection focus on discourse and social network analyses of online news platforms and fora \citep{warner2012detecting,becker2022decoding}, and social media platforms such as Twitter \citep{desmedt2021online,ron-etal-2023-factoring,chew2021quantifying,arviv2021echo,Chandra2021jewtocracy,mihaljević2023toxic,steffen2023codes,jikeli:axelrod:2022,ozalp,adl2018quantifying}, Facebook \cite{desmedt2021online,nefriana2025leader}, Reddit \citep{weinberg2025hidden}, YouTube \cite{barna2021exploration}, Telegram \cite{mihaljević2023toxic,steffen2023codes}, 4chan \cite{gonzalez:zannettou:23,ali2022analyzing}, and Gab \cite{bagavathi2019examining}.

To our knowledge, we are the first to use LLMs, or any computational method, for antisemitic event detection rather than hate speech detection. However, we build upon a small but growing body of research that examines various intersections between LLMs and antisemitism. Recent work has shown that commercial LLMs can be prompted to produce outputs containing both overt and covert forms of antisemitism, presenting a concerning safety risk of widely-deploying such technologies \citep{dutta2024down,gutmanmyth}. \citet{felkner2024gpt} present \textit{WinoSemitism}, a new benchmark dataset to evaluate such antisemitic biases in LLMs, containing stereotypical statements directly sourced from a survey of the Jewish community. Most relevant to our approach, LLMs such as ChatGPT and BERT have been evaluated on the task of antisemitic language detection \citep{mustafa2024can,patel2025evaluating}, but their performance is still far from perfect. Identifying antisemitic rhetoric is particularly challenging because it relies on historical knowledge, cultural stereotypes, and coded language such as dogwhistles \citep{mendelsohn2023dogwhistles}. Research on automatically identifying coded antisemitism remains nascent, but LLMs have also shown some potential for this task \citep{mendelsohn2023dogwhistles,kruk2024silent,kikkisetti2024coded,mustafa2024monitoring,sasse2025making}. Beyond linguistic analysis, recent work by \citet{rabbshort} highlights opportunities to reduce antisemitic attitudes via LLM-based conversational interventions.

\paragraph{Taxonomies of Hate}
Research on hate speech and abusive language detection has increasingly recognized the value of structured taxonomies with fine-grained (i.e. specific and detailed) labels over simple binary categories \citep{vidgen2021introducing,zufall-etal-2022-legal,khurana-etal-2022-hate}. We extend this approach to hateful event detection for several reasons. First, fine-grained taxonomies mitigate effects of annotator subjectivity. ``Harm'' is inherently subjective, and annotators often interpret it differently \citep{breitfeller-etal-2019-finding,sap-etal-2022-annotators,alkomah2022literature,kansok2023systematic,yin2021towards,fleisig-etal-2023-majority}. While broad labels such as ``harmful,'' ``toxic,'' or ``hateful'' frequently produce disagreement, hierarchical fine-grained classification can reduce subjectivity by specifying concrete forms of harm \citep{jiang2022investigating,xu2024leveraging,bibal-etal-2025-automating}. This is particularly important for antisemitism, where definitions are debated and interpretations of what actions should be considered as antisemitic vary widely \citep{Klug2023,Harrison:Klaff:2021,feldman2023antisemitism,herf:2021,penslar2022s,Nexus,jda}. Second, fine-grained taxonomies lead us to a more comprehensive understanding of LLM performance and the contexts in which models tend to make mistakes, thus informing better model development. The detailed labels themselves can also be useful analytic tools to characterize the specific forms of harm that Jewish people experience \citep{tripodi,Chandra2021jewtocracy}. Several annotation efforts unify conceptualizations of antisemitic rhetoric with expert-driven definitions and legal frameworks in order to develop fine-grained taxonomies tailored to antisemitism's unique historical and cultural context \citep{jikeli,becker2022decoding,becker2024discourse,becker2024guide}. By capturing the specific ways antisemitism manifests, fine-grained taxonomies enable more accurate monitoring and actionable insights.

\section{Datasets \& Taxonomies}

We compile four different datasets to evaluate large language models' ability to classify antisemitic incidents: \dataset{AMCHA}, \dataset{ADL-HEAT}, \dataset{Synthetic}, and \dataset{Campus-News}. The \dataset{AMCHA} and \dataset{ADL-HEAT} datasets were originally created by non-profit organizations dedicated to fighting antisemitism and hate more broadly. They contain descriptions and fine-grained labels of antisemitic incidents that occurred on higher education campuses and across the United States, respectively. Because \dataset{AMCHA} and \dataset{ADL-HEAT} only contain antisemitic incidents, they are alone insufficient for evaluating whether AI systems can distinguish reports of antisemitic from non-antisemitic incidents. We create and test models on the \dataset{Synthetic} dataset to address this limitation. Finally, we assess the generalizability and real-world utility of AI models on a new \dataset{Campus-News} dataset, containing real reports of both antisemitic incidents and non-antisemitic events. 

\subsection{AMCHA}

The AMCHA Initiative curates a database of English-language descriptions of antisemitic incidents that have occurred on higher education campuses, annotated for coarse- and fine-grained categories of antisemitism. This database provides uniquely rich content, labels, and contextual information (i.e., metadata) for hateful event classification.\footnote{An example of their general data collection methodology is described at \url{https://amchainitiative.org/wp-content/uploads/2023/12/Selective-Sympathy-Double-Standard-Report.pdf}.}

Domain experts from the AMCHA Initiative continuously monitor myriad sources, including lists of campus news publications, popular Jewish news publications, Google Alert keywords, antisemitism trackers on social media, anti-Zionist campus groups, and submissions from a reporting form. One team member then verifies that the news item covers an event that harmed Jewish people and impacted a higher education campus community. They further verify the factuality of the incident and the accuracy of the associated report. Another AMCHA Initiative team member then writes both a short and long description of the event. Depending on timing and the organization responsible for the event, the descriptor may customize a pre-built description template (templates were introduced in April 2024 to handle the rapidly increasing volume of events). The descriptions always conclude with links to the source(s) reporting the event as well as any available photo or video evidence linked to the event. The incident and its description are then tagged with labels from AMCHA's taxonomy. 

The two coarse-grained categories in \dataset{AMCHA} are \category{Targeting Jewish Students and Staff} (\category{``Targeting''}), which refers to incidents that directly target Jewish community members for harm based on their Jewishness or perceived association with Israel, and \category{Antisemitic Expression} (\category{``Expression''}), which refers to antisemitic language, imagery, or behavior. The seven fine-grained types within \category{Targeting} are: \type{Physical Assault}, \type{Discrimination}, \type{Destruction of Jewish property}, \type{Genocidal expression}, \type{Suppression of speech/movement/assembly}, \type{Bullying}, and \type{Denigration}. The two fine-grained types within \category{Expression} that we consider are: \type{Historical Antisemitism} and \type{Condoning Terrorism}. Each entry is labeled with one coarse-grained category and one or more fine-grained types.

We focus on clear-cut, relatively uncontroversial incidents of antisemitism (i.e. while there may be questions of which type of antisemitism an incident belongs to, the incident would be binarily classified as antisemitic according to both the IHRA and JDA definitions). We thus use the aforementioned types because both the International Holocaust Remembrance Alliance and Jerusalem Declaration on Antisemitism's definitions agree on their antisemitic nature. The AMCHA Initiative identifies three other \category{Expression} categories related to anti-Israel sentiment (\textit{Denying Jews Self-Determination}, \textit{Demonization of Israel}, and \textit{BDS Activity}). We omit entries that are exclusively labeled with anti-Israel types, but retain those that also contain labels for at least one type included in our study.

Our version of the \dataset{AMCHA} dataset contains 4,410 entries describing incidents over ten years, up to October 10, 2024. Each entry includes information about the date and location of the incident, as well as a 1-2 sentence natural language description. All \dataset{AMCHA} entries are sourced from publicly-viewable news and social media platforms. As an additional step to protect the privacy of individuals potentially named in the corpus, we use Microsoft's Presidio package,\footnote{\url{https://microsoft.github.io/presidio/text_anonymization/}} an anonymization tool that replaces people's names with a \texttt{<PERSON>} tag. We manually remove remaining names after running the tool. Table \ref{tab:amchastats} contains the set of categories and types analyzed, a brief description for each, and their frequency in our data.\footnote{The AMCHA Initiative's full descriptions: \url{https://amchainitiative.org/categories-antisemitic-activity} }

\begin{table}[htbp!]
\resizebox{\textwidth}{!}{%
\begin{tabular}{@{}llc@{}}
\toprule
\textbf{Coarse-Grained Category} & \textbf{Brief Description}                                                & \textbf{Frequency} \\ \midrule
\targeting                       & Directly targeting Jewish community members for harmful action            & 83.06\%            \\ 
\expression                      & Antisemitic language, imagery, or behavior                                & 16.94\%            \\ \midrule
\textbf{Fine-Grained Type}       & \textbf{Brief Description}                                                & \textbf{Frequency} \\ \midrule
\historical                      & Using symbols, images and tropes associated with historical antisemitism  & 33.06\%            \\
\condoning                       & Encouraging, justifying, or excusing the killing or harming of Jews       & 16.03\%            \\
\bullying                        & Tormenting Jewish community members because of their Jewishness           & 32.09\%            \\ 
\denigration                     & Unfairly ostracizing, vilifying or defaming Jewish community members      & 31.97\%            \\
\suppression                     & Suppression of Jewish students' speech, movement, and assembly            & 27.55\%            \\
\genocidal                       & Using imagery or language expressing a desire to eradicate Jews           & 22.02\%            \\
\destruction                     & Inflicting damage or destroying property owned by Jews or related to Jews & 8.89\%             \\
\discrimination                  & Unfair treatment or exclusion of Jewish community members                 & 7.41\%             \\
\physical                        & Physically attacking Jewish community members because of their Jewishness & 2.63\%             \\
\bottomrule
\end{tabular}%
}
\caption{Label descriptions and distributions for the \dataset{AMCHA} dataset containing 4,410 reports of antisemitic incidents. Note that fine-grained types are not mutually exclusive, so frequencies sum to more than 100\%.}
\label{tab:amchastats}
\end{table}

\subsection{ADL-HEAT}

The second dataset, \dataset{ADL-HEAT}, contains incident descriptions from the ADL H.E.A.T. (Hate, Extremism, Antisemitism, Terrorism) Map, a continuously-updated database of hateful and extremist incidents across the United States, focused on antisemitism, white supremacy, and anti-LGBTQ+ hate.\footnote{This information reflects the coverage of \dataset{ADL-HEAT} at the time of collection.} According to the ADL, the data ``is comprised of both criminal and non-criminal incidents of harassment, vandalism, and assault against individuals and groups as reported to ADL by victims, law enforcement, the media and partner organizations. It is not a public opinion poll or an effort to catalog every expression of antisemitism.''\footnote{ADL report with more details: \url{https://www.adl.org/resources/report/audit-antisemitic-incidents-2024}}. We download and filter the data collected as of December 25, 2024 to 4,522 incidents labeled with the ``Antisemitism'' tag. As with \dataset{AMCHA}, each entry in \dataset{ADL-HEAT} includes the date and location of the incident along with a 1-sentence description of the incident.

\begin{table}[ht!]
\resizebox{\textwidth}{!}{%
\begin{tabular}{@{}llc@{}}
\toprule
\textbf{Category} & \textbf{Brief Description}                                                & \textbf{Frequency} \\ \midrule
\harassment   & Verbal attacks on Jewish people, including slurs, stereotypes, tropes, or threats   & 62.84\%  \\
\vandalism    & Property damage accompanied by evidence of antisemitic intent or impact             & 35.01\%  \\ 
\assault      & Physical violence accompanied by evidence of antisemitic animus                     & 2.15\%   \\
\bottomrule
\end{tabular}%
}
\caption{Label descriptions and frequency for the \dataset{ADL-HEAT} dataset containing 4,522 antisemitic incident reports.}
\label{tab:adlstats}
\end{table}



\subsection{Synthetic}

\dataset{AMCHA} and \dataset{ADL-HEAT} are valuable for testing whether LLMs can perform fine-grained categorization of antisemitic incidents using short descriptions and basic contextual information. However, these datasets contain only incidents already judged to be antisemitic. They thus cannot reveal whether LLMs can effectively distinguish antisemitic from non-antisemitic incidents, a crucial capability for real-world deployment.

We address this limitation by creating the \dataset{Synthetic} dataset, containing LLM-generated descriptions of non-antisemitic events that relate to Jewish and/or Israeli people. We match the style and context of \dataset{AMCHA} and \dataset{ADL} as much as possible. We generate \dataset{Synthetic} through the following procedure:


\begin{enumerate}[itemsep=0pt]
    \item The first author, who has a background in studying and researching Jewish history and is ethnically Jewish, manually crafts a list of 12 phrases that describe positive (i.e., non-antisemitic) events related to Jewish and/or Israeli communities (e.g. ``Jewish folk dance class,'' ``Passover seder'').
    
    \item For each event, the author then manually and arbitrarily selects a reasonable date on which the event described could have occurred. 
    
    \item We specify a list of locations based on the existing datasets. To match \dataset{AMCHA}, we use locations from the AMCHA Initiative's list of 111 universities tracked for incidents. To match \dataset{ADL-HEAT}, we use the same set of 1,422 (city, state) pairs found in the database of antisemitic incidents. 
    
    \item Given the seed phrase describing an event, its corresponding date, and location, we use an LLM, OpenAI's \gpt, to generate a short report using the following prompt:

\begin{tcolorbox}[enhanced,breakable,width=\linewidth]
\begin{footnotesize}
\begin{Verbatim}[breaklines, breakanywhere,breaksymbol={}]
    Write a short (<300 tokens), objective news article about a {EVENT} that happened on {DATE} at {LOCATION}.
\end{Verbatim}
\end{footnotesize}
\end{tcolorbox}

Following setups for synthetic hate speech data generation task from prior work \citep{toxigen}, we set the temperature parameter to 0.9 for our generations to encourage higher diversity in the output text.

    \item We generate synthetic reports for all combinations of events and locations. We create 6 generations per combination for universities and 1 generation per combination for (city, state) pairs. This results in 7,992 reports matched to \dataset{AMCHA} ($6 \cdot 12 \cdot 111$) and 17,064 matched to \dataset{ADL} ($1 \cdot 12 \cdot 1422$).

    \item We take a random sample of this generated set to match the size of \dataset{AMCHA} and \dataset{ADL-HEAT} (4,410 and 4,522 entries, respectively).\footnote{All seed phrases for events, dates, and locations can be found at: \url{https://tinyurl.com/ASDetectionSyntheticValues}}
    
\end{enumerate}


We perform a manual inspection of 30 generated texts to ensure that they are coherent, relate to Jewish culture in some form, describe an event on the date and campus specified, and should be judged as benign events, and we find that all inspected texts pass these criteria. Nonetheless, we acknowledge that using synthetically generated texts carries risks of confounding factors in our experiments such as stylistic differences from real texts and value-match or construct-relevance errors. We thus caution that the results pertaining to \dataset{Synthetic} should be further validated.

\subsection{Campus-News}

Our final dataset, \dataset{Campus-News}, allows us to evaluate whether our best-performing AI system can identify and categorize antisemitic incidents in a more open-ended setting. This dataset reflects a real-world application in which models are applied on broad news coverage (rather than only content explicitly about Jews or antisemitism) to surface previously unseen antisemitic incidents. 

We scrape articles from campus newspapers for five of the ten universities with the most frequent incidents according to \dataset{AMCHA}: the Harvard Crimson,\footnote{\url{https://www.thecrimson.com/}} the Stanford Daily,\footnote{\url{https://stanforddaily.com/}} the Michigan Daily,\footnote{\url{https://www.michigandaily.com/}} the Daily Illini,\footnote{\url{https://dailyillini.com/}} and the Columbia Spectator.\footnote{\url{https://www.columbiaspectator.com/}} We collected a total of 5,275 articles from October 1, 2022 to December 25, 2024. Several annotators with expertise in antisemitism consensus-coded a sample of 225 articles according to the AMCHA taxonomy. Due to the time-consuming nature of reading and annotating long news articles, we selected items for manual annotation by first filtering the articles with a list of keywords related to Jews, Israel, Palestine, and Gaza. We then selected 225 of the most recent articles, roughly evenly distributed across publications. We note that the small sample size of manually annotated articles limits the statistical power of our conclusions; we nonetheless provide preliminary insights for mapping the landscape of antisemitic events on college campuses and for potentially using AI to scale datasets such as \dataset{AMCHA}.

\section{Methods}
\subsection{Event Classification}
\begin{table}[htbp!]
\resizebox{\textwidth}{!}{%
\begin{tabular}{c|ccc}
\toprule
\textbf{Variation} & \textbf{Assumes Event is Antisemitic} & \textbf{Defines Terms}                                                & \textbf{Contains In-Context Examples} \\ \midrule
\unforcednodef & No & No & No \\
\forcednodef & Yes & No & No \\
\unforceddef & No & Yes & No \\
\forceddef & Yes & Yes & No \\
\oneshot & Yes & No & Yes \\
\bottomrule
\end{tabular}%
}
\caption{Reference table of prompt variations for event classification.}
\label{tab:methods}
\end{table}
We conduct antisemitic incident classification using two generative LLMs: OpenAI's closed-source \gpt and Meta's open-weight \texttt{Llama-3.2-3b-Instruct} (\llama), which can be downloaded and run locally. For our experiments, we focus on prompting rather than fine-tuning or task-specific classifiers because it is most financially and logistically practical for the intended users of this antisemitism classification tool. Furthermore, the size of the dataset is relatively small (especially broken down per category/type), constituting a likely insufficient amount of data to fine-tune or train a task-specific classifier from scratch.

For both models, we systematically evaluate multiple prompting strategies. In our baseline setup, we formulate a prompt that includes the description, location, and time of occurrence for a given incident. After the incident information, the prompt requests that the LLM provide labels for (1) whether or not the text describes an antisemitic event and (2) the most appropriate coarse-grained category. For entries annotated according with the AMCHA taxonomy, the prompt additionally requests (3) all applicable fine-grained types. We call this baseline setup \unforcednodef because it provides no other contextual information to guide the LLM. We consider four additional prompt variations: 

\begin{enumerate}[topsep=0pt,noitemsep]
    \item \forcednodef: Asking the LLM to presuppose or \textbf{assume} that the incident described is indeed antisemitic and respond accordingly. In this setting, we remove the binary antisemitic-or-not task and only prompt the model for coarse-grained categories and fine-grained types when applicable.

    \item \unforceddef: Including a \textbf{definition} of antisemitism and each label. Specifically, we begin the prompt with Wikipedia's general definition of antisemitism (corroborated by the IHRA's definition): \textit{Antisemitism is defined as hostility to, prejudice towards, or discrimination against Jews}. Rather than just providing names of possible categories and types in the prompt, we also provide a brief explanation of each label in the \unforceddef setting.

    \item \forceddef: Combining the above two variations, both adding definitions of each label and asking the LLM to presuppose binary antisemitism.

    \item \ice: Including \textbf{in-context examples} in the prompt to guide the LLM's outputs. We include one randomly-selected labeled example for each possible fine-grained type (or coarse-grained category for \dataset{ADL-HEAT}). This approach is also known as \textit{in-context learning} or \textit{few-shot learning}. In this setting, we also ask the LLM to presuppose binary antisemitism.

\end{enumerate}
We visualize all five prompt variations in Table \ref{tab:methods}. Below is the exact text for the prompt used in the \unforceddef setting. We use the minimalistic Wikipedia definition of antisemitism for this prompt to avoid prejudging any contested categories and types related to Israel. Appendix \ref{subsec:prompts} lists the prompt templates for the other four variations. All prompt variations are available in our code repository.\footnote{\url{https://github.com/ENSCMA2/antisemitic-incident-classification}}

\begin{tcolorbox}[enhanced,breakable,width=\linewidth]
\begin{footnotesize}
\begin{Verbatim}[breaklines, breakanywhere,breaksymbol={}]
Antisemitism is defined as hostility to, prejudice towards, or discrimination against Jews. The following text, enclosed in single quotes, describes an event that happened on 2024-11-07 00:29:00 at Stanford University.

'[article text goes here]'

Does this quoted text describe an antisemitic event? If so, what category of antisemitism does the event fall under, and what type of expression does it fall under? Give an answer in the following format:

{'describes_antisemitic_event': bool,
'category': char,
'type_of_expression': int list}

For 'category', put the letter corresponding to the strongest option below for the quoted text:
A: 'Does not describe an antisemitic incident'
B: 'antisemitic expression' - Language, imagery or behavior deemed antisemitic by the U.S. State Department definition of antisemitism, or wholly consistent with that definition
C: 'targeting Jewish students and staff' - Incidents that directly target Jewish students on campus or other Jewish members of the campus community for harmful or hateful action based on their Jewishness or perceived support for Israel

For 'type_of_expression', list all numbers corresponding to the the following options that apply to the quoted text:
0: 'Does not describe an antisemitic incident'
1: 'physical assault' - Physically attacking Jewish students or staff because of their Jewishness or perceived association with Israel
2: 'discrimination' - Unfair treatment or exclusion of Jewish students or staff because of their Jewishness or perceived association with Israel
3: 'destruction of Jewish property' - Inflicting damage or destroying property owned by Jews or related to Jews
4: 'genocidal expression' - Using imagery (e.g. swastika) or language that expresses a desire or will to kill Jews or exterminate the Jewish people
5: 'suppression of speech/movement/assembly' - Preventing or impeding the expression of Jewish students, such as by removing or defacing Jewish students' flyers, attempting to disrupt or shut down speakers at Jewish or pro-Israel events, or blocking access to Jewish or pro-Israel student events
6: 'bullying' - Tormenting Jewish students or staff because of their Jewishness or perceived association with Israel
7: 'denigration' - Unfairly ostracizing, vilifying or defaming Jewish students or staff because of their Jewishness or perceived association with Israel
8: 'historical' - Using symbols, images and tropes associated with historical antisemitism, including by making "mendacious, dehumanizing, demonizing, or stereotypical allegations about Jews as such, or the power of Jews as a 
collective-especially but not exclusively, the myth about a world Jewish conspiracy or of Jews controlling the media, economy, governments, or other societal institutions"
9: 'condoning terrorism' - Calling for, aiding or justifying the killing or harming of Jews
\end{Verbatim}
\end{footnotesize}
\end{tcolorbox}

\subsection{Model Evaluation}

We compute the \textit{binary detection rate}: how often the model predicts that the text describes an antisemitic incident. The true binary detection rate for \dataset{AMCHA} and \dataset{ADL-HEAT} is 100\% since all of the entries in these datasets describe antisemitic events. On the contrary, the true binary detection rate for \dataset{Synthetic} is 0\%, as this dataset was specifically constructed to assess performance on non-antisemitic events. 

We formulate \textbf{coarse-grained category} prediction as a multi-class classification task, where each document is assigned to exactly one category. By contrast, \textbf{fine-grained type} prediction is a multi-label classification task, where each document may be assigned to multiple types. While we report model performance separately for these two tasks, they share the same evaluation procedure.

Evaluation metrics compare the true (``gold'') labels for each document with model predictions, and include: \textit{accuracy} (how often the correct label is predicted), \textit{precision} (how often predicted positive labels are correct), and \textit{recall} (how often gold positive labels are correctly predicted by the model). We additionally compute F1 scores, which are a standard machine learning evaluation metric that balances precision and recall.

Consider a class $i$, which can either be from the set of categories or types. Let $N_i$ and $\hat{N}_i$ be the number of documents labeled with class $i$ in the gold and predicted data, respectively. $TP_i$ is the number of \textit{true positives} (when the model correctly assigns class $i$ to the document). The per-class F1 score ($F1_i$) and the weighted average over all classes ($WF1$) are calculated as:

$$F1_i = \frac{2 \cdot TP_i}{N_i + \hat{N}_i} \qquad\qquad WF1 = \frac{\sum_{i}N_i \cdot F1_i}{\sum_{i}N_i}$$

To aid interpretation, we also group types into \textbf{action} (primarily involving physical actions) and \textbf{rhetoric} (primarily involving verbal expressions of hate). In reality, the boundaries between actions and rhetoric are fluid; for example, drawing swastika graffiti (\genocidal) is both an expression and an action, and \discrimination can occur through both linguistic and physical mechanisms. However, we believe this framing is useful because it represents the key distinction between the more familiar task of hate \textit{speech} detection (primarily rhetoric) and hateful \textit{event understanding} (events may involve more actions).

\section{Results}
We present results on \dataset{AMCHA} and \dataset{ADL-HEAT} in order of increasing task granularity: binary antisemitic incident detection, coarse-grained category assignment, and fine-grained type classification. The section concludes with results from \dataset{Campus-News}, the real-world case study.

\subsection{Binary Antisemitic Incident Detection}

Table \ref{tab:binary} shows the binary detection rate for each LLM using the baseline \unforcednodef prompt and the \unforceddef prompt that includes definitions. This is the only task that includes the \dataset{Synthetic} dataset. It excludes \forcednodef prompt variations, which do not ask the LLM to make the binary antisemitic-or-not judgment. 

Overall, both models perform quite well on the binary antisemitism classification task. The \llama model tends to label incidents as antisemitic more frequently than \gpt, leading to both higher true positive and false positive rates. Adding definitions has an inconsistent effect on model performance. Definitions make \llama more likely to label incidents as antisemitic, but this effect is small for true positives (\dataset{AMCHA} and \dataset{ADL-HEAT}) and larger for false positives (\dataset{Synthetic}). In contrast, adding definitions substantially \textit{decreases} detection of true positives for \gpt on \dataset{ADL-HEAT}.

\begin{table}[htbp!]
\centering
\resizebox{.6\textwidth}{!}{%
\begin{tabular}{@{}ccccc@{}}
\toprule
                    & \multicolumn{2}{c}{\gpt} & \multicolumn{2}{c}{\texttt{Llama-3.2-3b-Instruct}} \\ 
                    & \unforcednodef        & \unforceddef         & \unforcednodef       & \unforceddef         \\ \midrule
\dataset{AMCHA}     & 93.58\%      & 93.76\%     & 99.71\%     & 99.91\%     \\
\dataset{ADL-HEAT}  & 97.63\%      & 90.78\%     & 98.39\%     & 98.45\%     \\ \midrule
\dataset{Synthetic} & 0\%          & 0\%         & 1.56\%      & 13.36\%     \\ \bottomrule
\end{tabular}%
}
\caption{Percentage of entries labeled as antisemitic by each model and prompt combination. Ground truth for \dataset{AMCHA} and \dataset{ADL-HEAT} is 100\% antisemitic, while ground truth for \dataset{Synthetic} is 0\% antisemitic.  We observe that \gpt returns no false positives from \dataset{Synthetic}, while some more confusion occurs with \llama. However, \gpt has more false negatives than \llama, though both models achieve under 10\% false negative rates. Although we manually inspected a sample of our synthetic data and believe it provides useful signals of performance, we caution that due to the nature of synthetic texts in \dataset{Synthetic}, the bottom row of results can benefit from further validation.}
\label{tab:binary}
\end{table}

\subsection{Coarse-Grained Category Assignment}

We now examine how each model and prompt variation performs on correctly assigning incidents to coarse-grained categories: \targeting and \expression for \dataset{AMCHA}, and \harassment, \vandalism, and \assault for \dataset{ADL-HEAT}.  

\begin{figure}[t]
\centering

\begin{minipage}[t]{0.58\linewidth}
  \vspace{0pt}%
  \centering
  \includegraphics[
    trim={2.8em 4em 4em 4em},
    clip,
    width=\linewidth
  ]{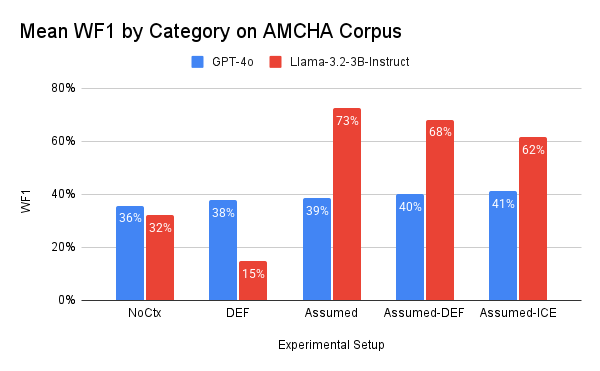}
  \captionof{figure}{
    Weighted F1 scores for coarse-grained category classification on \dataset{AMCHA}. Results are distinguished by model (bar color) and prompt variation (x-axis). The best-performing model/variation combination is \forcednodef on \llama. The binary assumption of antisemitism, term definitions (along with the binary assumption), and in-context examples all help with performance compared to the \unforcednodef baselines for both models.
  }
  \label{fig:amcha-coarse}
\end{minipage}\hfill
\begin{minipage}[t]{0.37\linewidth}
  \vspace{5mm}%
  \centering
  \resizebox{\linewidth}{!}{%
    \begin{tabular}{@{}llcc@{}}
      \toprule
       & & \targeting & \expression \\ \midrule
      \gpt & \unforcednodef & 36\% & 36\% \\
                      & \unforceddef   & 38\% & 35\% \\
                      & \forcednodef   & 39\% & 35\% \\
                      & \forceddef     & 41\% & 36\% \\
                      & \oneshot       & 43\% & 36\% \\ \midrule
      \llama  & \unforcednodef & 32\% & 32\% \\
                      & \unforceddef   & 12\% & 30\% \\
                      & \forcednodef   & 81\% & 34\% \\
                      & \forceddef     & 75\% & 36\% \\
                      & \oneshot       & 67\% & 35\% \\ \bottomrule
    \end{tabular}
  }
  \captionof{table}{Per-category F1 scores for each model and prompt variation on \dataset{AMCHA}. Scores are relatively low across the board but exhibit improvement, especially with the \targeting category, with binary assumptions, in-context examples, and term definitions.}
  \label{tab:amcha-coarse}
\end{minipage}

\end{figure}

\paragraph{\dataset{AMCHA}} 
\gpt has consistently low performance in classifying \targeting and \expression categories across prompts (Figure \ref{fig:amcha-coarse}). Weighted F1 scores range from 36\% with the baseline \unforcednodef prompt to 41\% with \oneshot prompt. Adding definitions, instructing the model to assume the incident is antisemitic, and providing in-context examples all improve \gpt's performance in identifying instances of targeting Jewish students and staff (\targeting), though effects are small (Table \ref{tab:amcha-coarse}). However, these variations do not improve the model's ability to identify antisemitic expression. 

Prompt variations affect \llama in very different ways. Assuming the incident is antisemitic (\forcednodef) dramatically improves performance, increasing the weighted F1 score from 32\% to 73\%. On the other hand, including definitions (\unforceddef) decreases weighted F1 from 32\% in \unforcednodef to 15\% in \unforceddef, and from 73\% in \forcednodef to 68\% in \forceddef. In-context examples also hurt performance for \llama, decreasing weighted F1 from 73\% in \forcednodef to 62\% in \oneshot. As with \gpt, the effects on weighted F1 are primarily driven by changes in the per-category F1 score for \targeting rather than \expression.

\paragraph{\dataset{ADL-HEAT}}

\begin{figure}[t]
\centering

\begin{minipage}[t]{0.52\linewidth}
  \vspace{0pt}%
  \centering
  \includegraphics[
    trim={1em 4em 4em 0em},
    clip,
    width=\linewidth
  ]{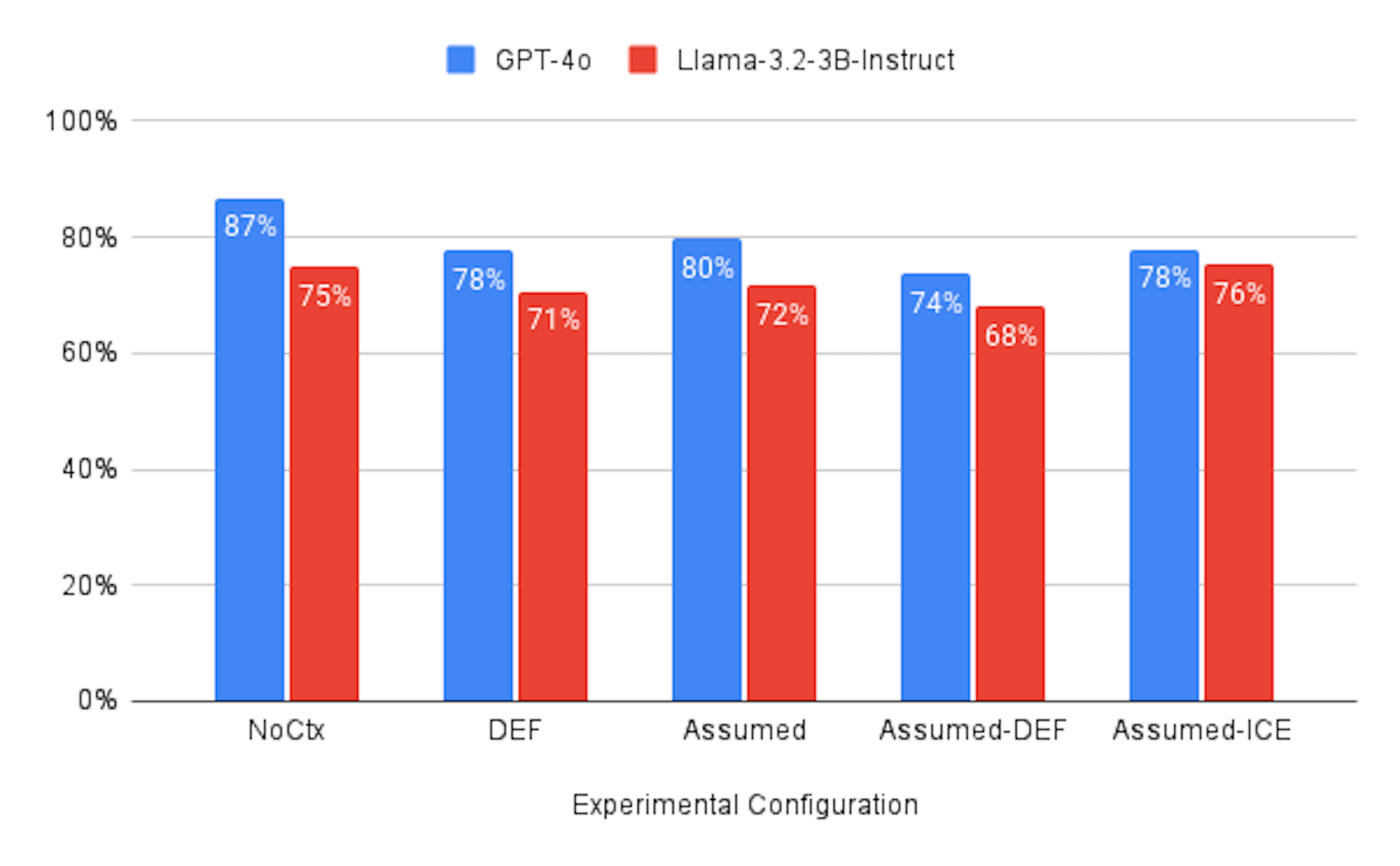}
  \captionof{figure}{
    Weighted F1 scores across coarse-grained categories for the
    \dataset{ADL-HEAT} dataset. Results are distinguished by model (bar color) and prompt variation (x-axis). \gpt performs best on this task under the \unforcednodef variation (no term definitions or in-context examples, outperforming all \llama settings as well), while the best setting for \llama is \oneshot, indicating that \llama benefits slightly from in-context examples and the binary assumption of antisemitism.
  }
  \label{fig:adl-coarse}
\end{minipage}\hfill
\begin{minipage}[t]{0.45\linewidth}
  \vspace{5mm}%
  \centering
  \resizebox{\linewidth}{!}{%
    \begin{tabular}{@{}llccc@{}}
\toprule
       &                        & \assault & \harassment & \vandalism \\ \midrule
\gpt & \unforcednodef       & 88\%    & 85\%       & 91\%      \\
       & \unforceddef         & 89\%    & 81\%       & 72\%      \\
       & \forcednodef       & 88\%    & 73\%       & 91\%      \\
       & \forceddef         & 89\%    & 73\%       & 75\%      \\
       & \oneshot              & 84\%    & 79\%       & 77\%      \\ \midrule
\llama  & \unforcednodef       & 18\%    & 81\%       & 68\%      \\
       & \unforceddef         & 42\%    & 76\%       & 64\%      \\
       & \forcednodef        & 46\%    & 74\%       & 69\%      \\
       & \forceddef         & 60\%    & 72\%       & 61\%      \\
       & \oneshot           & 13\%    & 79\%       & 73\%      \\ \bottomrule
\end{tabular}%
}
  \captionof{table}{Per-category F1 scores for each model and prompt variation on \dataset{ADL-HEAT}. \gpt performs better on the task overall, with \assault having the highest F1 scores and term definitions appearing to contribute the most performance improvements.}
  \label{tab:adl-coarse}
\end{minipage}

\end{figure}

Overall, both models are better at labeling categories from \dataset{ADL-HEAT}, with F1 scores ranging from 68\% (\llama; \forceddef) to 87\% (\gpt; \unforcednodef) (see Figure \ref{fig:adl-coarse}). \gpt outperforms \llama for all prompt variations. In contrast to what we observe with \dataset{AMCHA}, adding definitions and asking the model to assume an incident is antisemitic harm model performance for both LLMs. Adding in-context examples slightly helps \llama but not \gpt. 

More similarities between the two LLMs emerge when focusing on per-category F1 scores (Table \ref{tab:adl-coarse}). For \harassment, both models have the best performance with the baseline \unforcednodef prompt. Asking the model to assume antisemitism substantially decreases scores, as does including definitions, though to a lesser extent. The performance drop from \forcednodef is somewhat mitigated by including in-context examples (\oneshot). For \vandalism, both \gpt and \llama do best with the \unforcednodef and \forcednodef prompt variations, with big drops in performance when adding definitions and in-context examples. In-context examples also hurt both models' F1 scores for the \assault category. In contrast to the observed patterns for the other categories, definitions and assuming antisemitism drastically help \llama correctly label instances of \assault.

Our experiments show that LLMs tend to classify categories from \dataset{ADL-HEAT} more accurately than those in \dataset{AMCHA}, but prompt augmentations are more helpful for \dataset{AMCHA}. This task corroborates prior work suggesting that there is no universally optimal prompt formulation \citep{atreja2025s}. Rather, the best prompting strategy is dependent on the choice of LLM and dataset.

\begin{figure}[htbp!]
    \centering
    \includegraphics[
    trim={4em 4em 4em 0em},
    clip,
    width=.6\linewidth]{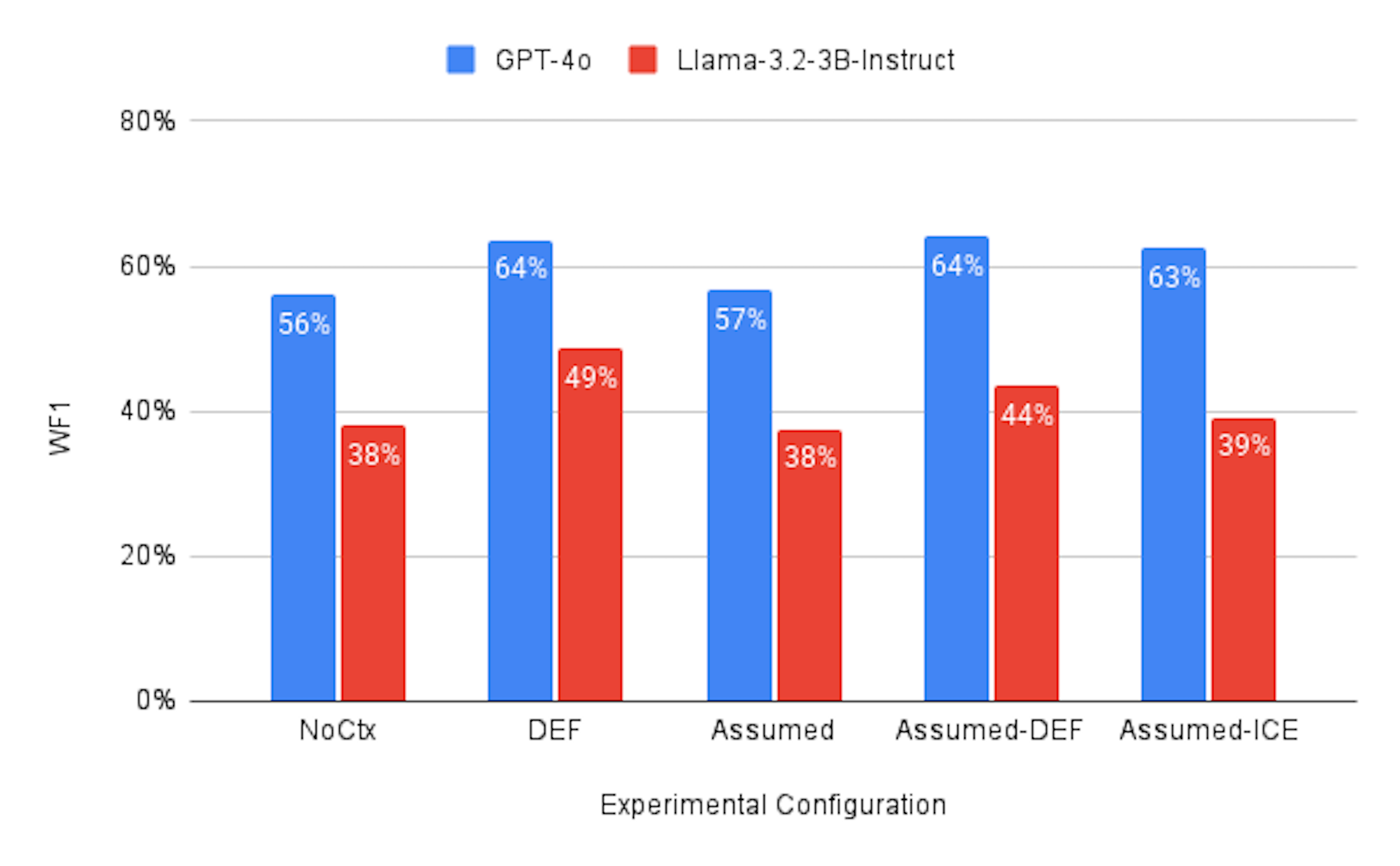}
    \caption{Weighted F1 scores across fine-grained types for the \dataset{AMCHA} dataset. Results are distinguished by model (bar color) and prompt variation (x-axis). Term definitions provide the most improvement for both models, with \gpt performing best overall.}
    \label{fig:amcha-fine}
\end{figure}

\subsection{Fine-Grained Type Classification}

We now evaluate LLMs on fine-grained type classification using the \dataset{AMCHA} dataset. As Figure \ref{fig:amcha-fine} illustrates, \gpt outperforms \llama across all prompt variations by a large margin. Surprisingly, \gpt even labels fine-grained types much more accurately than coarse-grained categories. For both LLMs, including definitions and examples in the prompt improves overall weighted F1 scores. 

We further analyze the per-type F1 scores from \gpt (Table \ref{tab:amcha-fine}). There is substantial variation in model performance across fine-grained types. \gpt works very well for labeling incidents as \condoning or \physical, with per-type F1 scores above 80\% for all prompt variations. On the other hand, F1 scores are mostly below 50\% for \bullying and \discrimination.

\begin{figure}[htbp!]
    \centering
    \includegraphics[
    trim={0em 0em 4em 0em},
    clip,
    width=.6\linewidth]{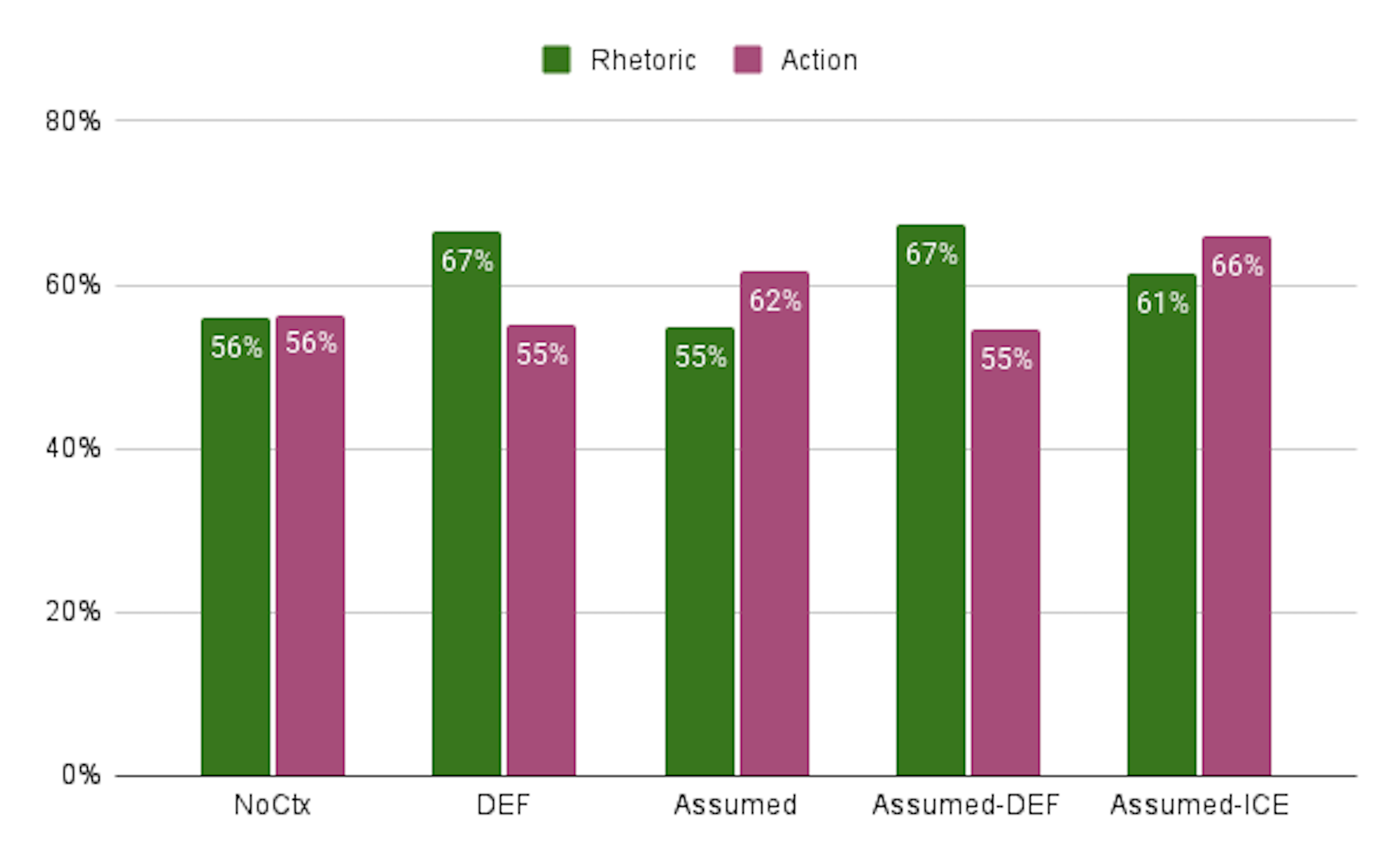}
    \caption{Weighted F1 scores across rhetoric and action-oriented fine-grained types for the \dataset{AMCHA} dataset for all prompt variations. Term definitions appear to improve performance on rhetoric-related types, while the binary assumption of antisemitism and some in-context examples help improve performance on action-related types.}
    \label{fig:rhetoric-action}
\end{figure}

Figure \ref{fig:rhetoric-action} compares weighted F1 scores for rhetoric-oriented versus action-oriented types. In aggregate, \gpt achieves comparable performance across rhetoric and action types, but the effects of prompt augmentations differ systematically between them. For rhetoric types, including definitions (\unforceddef and \forceddef) yields the strongest performance gains with weighted F1 increasing from 56\% to 67\%. For action types, including correctly-labeled example incidents (\oneshot) boosts performance the most, from 56\% to 66\% weighted F1. In contrast to rhetoric types, definitions do not help---and even slightly hurt---model performance for action types.  One notable partial exception to this pattern is \discrimination: both definitions and examples substantially improve scores, possibly because \discrimination can manifest as both rhetoric and action.

\begin{table}[htbp!]
\resizebox{\textwidth}{!}{%
\begin{tabular}{@{}cccccc|cccc@{}}
\toprule
& \multicolumn{5}{c|}{Rhetoric}   & \multicolumn{4}{c}{Action}  \\ 
&\condoning &\historical &\genocidal &\denigration &\bullying &\physical &\discrimination  &\suppression &\destruction \\\midrule
\unforcednodef  & 84.5\% & 61.8\% & 57.2\% & 52.9\% & 38.0\% & 86.8\% & 42.8\% & 53.9\% & 66.5\% \\
\unforceddef & 81.1\% & \textbf{74.7\%} & 73.6\%  & 63.1\% & 49.8\% & 87.6\% & 48.5\% & 51.2\% & 63.0\% \\
\forcednodef & \textbf{86.2\%} & 52.6\% & 55.5\%  & 55.6\% & 40.8\% & 87.3\% & 45.0\% & 62.0\% & 66.8\% \\
\forceddef &82.5\%&73.2\% &\textbf{78.0\%}&\textbf{64.1\%}&\textbf{49.9\%}&87.4\% &\textbf{50.5\%}&49.2\%&64.7\%\\
\oneshot &85.5\% &64.8\%&63.2\% &62.9\%&43.1\% &\textbf{88.0\%}&48.6\%&\textbf{65.9\%}&\textbf{73.6\%} \\\bottomrule
\end{tabular}%
}
\caption{Per-type F1 scores for \gpt on \dataset{AMCHA}, separated by prompt variation. The highest scores per type are shown in \textbf{bold}. \forceddef performs best for three rhetoric-related types, while in-context examples help with three action-related types.}
\label{tab:amcha-fine}
\end{table}

\paragraph{Role of cultural knowledge}
Correctly identifying some types, particularly \historical and \genocidal, requires substantial historical and cultural contextual knowledge. The \historical category captures forms of classical antisemitism, such as conspiracy, power, dehumanization, and dual loyalty tropes; these are often expressed indirectly and require familiarity with longstanding antisemitic narratives. Similarly, \genocidal includes expressions that call for or endorse the genocide of Jewish people. While such calls may be overt, they often manifest through implicit or coded references to the Holocaust. Though we thus initially expected \gpt to struggle with these types, weighted F1 scores for \historical and \genocidal are comparable to other types (Table \ref{tab:amcha-fine}). Out of all types, \historical and \genocidal benefit the most from including definitions, suggesting that this additional conceptual grounding is particularly important for less overt forms of hateful incidents.

One possible explanation for \gpt's superior performance on \historical and \genocidal compared to \bullying and \discrimination is that \historical and \genocidal constitute more stable and globally-recognized types. They draw upon well-established recurring narratives that may be widely represented in the massive datasets used for training LLMs. On the other hand, \bullying and \discrimination hinge on local, situational, and interpersonal context. Discerning these types requires reasoning about intent and social relations that may not be explicit in event descriptions nor LLM training data.

\paragraph{Error analysis}
Table \ref{tab:misclass} shows several examples of model errors at both the coarse-grained category and fine-grained type levels. In the first example, most models missed that the incident describes a covert expression of \historical tropes through the statement ``Ye was Right.'' While the statement does not directly evoke any tropes, it expresses agreement with Kanye West's remarks, which heavily draw upon \historical antisemitism. In the second example, several models missed that the online threats constitute both \bullying and \suppression of Jewish students' right to movement and assembly.

For the third example, most models correctly identified that swastika graffiti is both \historical and \genocidal. \dataset{AMCHA} considers such calls or affirmations of genocide to be a form of \targeting, but all models categorize this incident as \expression. This discrepancy may not truly be a model error, but rather an artifact of the taxonomy. A swastika is indeed a form of antisemitic expression because it is a form of antisemitic imagery, so one could argue that \expression is more directly applicable than \targeting. However, \dataset{AMCHA}'s creators opted to make \targeting and \expression mutually-exclusive categories, and label incidents solely as \targeting if both apply.

\begin{table*}[htbp!]
    \footnotesize
    \centering
    \begin{tabular}{@{}p{7.2cm}p{1.75cm}p{1.75cm}p{1.75cm}p{1.75cm}@{}}\toprule
       \textbf{Text} & \textbf{Gold} & \unforcednodef & \oneshot & \unforceddef\\\midrule
       Chalking stating ``Ye was Right,'' which referenced antisemitic comments made by the rapper <PERSON>, and ``It's Not Cool to Shill for Israel'' was found on Bruin Walk.     & \expression; \historical & \expression; \denigration & \expression; \denigration & \expression; \denigration, \historical \\\midrule
      According to the ADL, a University at Buffalo student made online threats against an on-campus march organized by the school's Jewish Student Union. & \targeting; \bullying, \suppression & \targeting; \bullying, \suppression & \targeting; \bullying & \targeting; \bullying \\\midrule
      Swastika graffiti was found on a fence post. & \targeting; \genocidal, \historical & \expression; \denigration, \historical & \expression; \genocidal, \historical & \expression; \genocidal, \historical
     \\\bottomrule
    \end{tabular}
    \caption{Examples of classification errors on entries from the AMCHA Corpus. Errors can occur at the coarse-grained category level (bolded) or the fine-grained type level (italicized).}
    \label{tab:misclass}
\end{table*}

\subsection{Campus-News Case Study}

We select the best-performing setup on fine-grained type detection, \gpt with the \unforceddef prompt, and apply it to \dataset{Campus-News}. We first compare the LLM labels with the human annotations for 224 articles. Reports of antisemitic incidents are quite sparse in the university newspapers, with only 19 articles (8.5\%) describing such an incident. The small number of positive examples per label yields highly noisy performance estimates, which should thus be interpreted with caution. Across nearly all categories and types, the LLM achieves considerably higher recall than precision. In other words, false positives are much more common than false negatives; the model tends to be assign antisemitic (and sub-type) labels more liberally than human annotations. The one exception is \targeting, which has higher precision than recall. This is likely a result of how \dataset{AMCHA} constructed this category, where all incidents containing both \expression and \targeting elements are assigned to solely the latter.

\begin{table}[htbp!]
\centering
\resizebox{.6\textwidth}{!}{%
\begin{tabular}{@{}lcccc@{}}
\toprule
Label          & Count & Precision & Recall & F1    \\ \midrule
\textit{\textbf{Binary}} & 19    & 27.0\%     & 52.6\%  & 35.7\% \\ \midrule
\expression     & 4     & 8.0\%       & 50.0\%   & 13.8\% \\
\targeting      & 15    & 33.3\%     & 26.7\%  & 29.6\% \\ \midrule
\condoning      & 5     & 4.2\%      & 60.0\%   & 7.9\%  \\ 
\historical     & 1     & 0.0\%       & 0.0\%    & 0.0\%   \\
\genocidal      & 2     & 2.3\%      & 100\%  & 4.4\%  \\
\denigration    & 3     & 0.0\%      & 0.0\%    & 0.0\%   \\
\bullying       & 5     & 1.6\%      & 20.0\%   & 3.0\%  \\
\physical       & 0     & N/A       & N/A    & N/A   \\
\discrimination & 0     & N/A       & N/A    & N/A   \\
\suppression    & 7     & 2.4\%      & 28.6\%  & 4.5\%  \\
\destruction    & 2     & 1.3\%      & 50.0\%   & 2.6\%  \\ \bottomrule
\end{tabular}%
}
\caption{Model performance on the human-annotated portion of the \dataset{Campus-News} dataset using \gpt with the \unforceddef prompt. \textit{Count} refers to the number of entries assigned each label in the human-provided annotation. Overall, we observe moderate recall with lower precision, though the pattern is flipped for \targeting.}
\label{tab:news-performance}
\end{table}

\paragraph{Surfacing Antisemitic Incidents on Campuses.}

Model performance results on \dataset{Campus-News} suggest that LLMs may be useful as an initial filter to flag potentially antisemitic incidents for further review by human experts.
We demonstrate the potential for our method for surfacing antisemitic incident reports on the full \dataset{Campus-News} dataset, but caution that results are highly preliminary and should be further validated.

Out of 5,275 news articles containing keywords related to Jews, Israel or Palestine, 364 (6.9\%) are predicted to describe antisemitic incidents. The relative proportion of antisemitic incident reports varies from just 0.9\% (for the Daily Illini) to 13.8\% (for the Columbia Spectator). In contrast to \dataset{AMCHA} and the sample of \dataset{Campus-News} annotated by humans, the model assigns \expression more frequently than \targeting overall. However, the predicted distribution of \expression vs. \targeting also varies across publications: while reports of antisemitic incidents in the Columbia Spectator are nearly evenly split between \expression and \targeting, 80.6\% of those in the Michigan Daily labeled as \expression.

Quantifying the nature of antisemitic incidents has implications for informing intervention strategies in higher education settings. \targeting incidents indicate hostile actions towards Jewish students and staff, and would suggest the need for more on-the-ground interventions to ensure Jewish safety on campus, while \expression incidents point to a greater need to dedicate resources for campus-wide anti-hate education. Qualitatively, we also observe that flagged incidents from campus newspapers are often part of broader stories with national relevance, for example by covering high-profile lawsuits, Congressional hearings, and FBI reports. With our small sample of articles, we begin to see that surfaced reports could help practitioners understand the local impacts of such prominent stories, and help identify local organizations to work with for combating antisemitism.

\begin{table}[]
\centering
\resizebox{.7\textwidth}{!}{%
\begin{tabular}{@{}lcccc@{}}
\toprule
Publication & \# Articles & \begin{tabular}[c]{@{}l@{}}\# Predicted \\Antisemitic\\Incidents\end{tabular}
 & \% \expression & \% \targeting \\ \midrule
Harvard Crimson    & 1716 & 139 (8.1\%)  & 67.6\% & 32.4\% \\
Stanford Daily     & 950  & 43 (4.5\%)   & 60.5\% & 39.5\% \\
Daily Illini       & 224  & 2 (0.9\%)    & 50.0\% & 50.0\% \\
Michigan Daily     & 1307 & 31 (2.4\%)   & 80.6\% & 19.4\% \\
Columbia Spectator & 1078 & 149 (13.8\%) & 49.7\% & 50.3\% \\ \midrule
Total              & 5275 & 364 (6.9\%)  & 60.4\% & 39.6\% \\ \bottomrule
\end{tabular}%
}
\caption{Frequency of \dataset{Campus-News} articles predicted by the LLM (\gpt with \unforceddef prompt) to describe an antisemitic incident, across five campus publications. Among surfaced articles, the distribution of \expression and \targeting categories are shown. We see relatively high frequencies of \expression-related incidents, with lower but still frequent incidents constituting \targeting.}
\label{tab:news-surfacing}
\end{table}

\section{Discussion}

This work introduces the novel task of \textit{fine-grained hateful event detection}, designed to capture not only whether a hateful event occurred, but also the specific form of harm. Distinct from hate speech detection, this task bridges online rhetoric and offline events, drawing on news articles, civil society reports, and official records as data sources. Event reports provide concrete details about how marginalized communities are harmed. Processing these reports in a structured and efficient manner facilitates timely monitoring of hateful incidents and trends, which can further inform social scientific research as well as counter-hate policies and interventions.

We study this task in the context of antisemitism by compiling multiple datasets of event descriptions: \dataset{AMCHA} (antisemitic incidents in U.S. higher education settings with annotations for coarse-grained categories and fine-grained types), \dataset{ADL-HEAT} (antisemitic incidents across the U.S. labeled as harassment, assault, and vandalism), and \dataset{Synthetic} (non-antisemitic but Jewish-related event descriptions to test for false positives). We evaluate two state-of-the-art LLMs, \gpt and \longllama on antisemitic incident detection and document the impact of various prompting strategies, including adding label definitions, in-context examples, and a binary assumption of antisemitism. We further assess generalization by applying our best-performing setup to recently-collected articles from college newspapers containing previously unseen reports.

Several distinct patterns emerge from our evaluation of LLMs on antisemitic incident detection. First, performance is highly contingent on the specific choice of taxonomy and task: almost every model and prompt combination classifies antisemitic incidents from \dataset{ADL-HEAT} more accurately than \dataset{AMCHA}. This is likely due to task difficulty: \dataset{ADL-HEAT} requires models to choose between \harassment, \vandalism, and \assault, which are all quite conceptually distinct. On the other hand, \dataset{AMCHA} requires models to choose between the more abstract categories of \targeting and \expression, and the differences between fine-grained labels such as \denigration and \bullying are more nuanced. The difference in task difficulty could also explain the diverging effects of prompt augmentations between the two datasets: including definitions and in-context examples tends to help for \dataset{AMCHA}, where additional context may be necessary. However, adding extra information to the prompt hurts for \dataset{ADL-HEAT}, where the drawbacks of increasing prompt length and complexity outweigh possible informational gains. Beyond task difficulty, there may be yet-unexplored effects of original data sources, event description styles, and annotation procedures on model performance.

Second, \gpt generally outperforms \llama. While \llama has the highest performance on coarse-grained incident classification for \dataset{AMCHA}, its performance varies drastically across different prompt variations; such noisy estimates may be due to its much smaller size compared to \gpt (while the exact number of parameters in \gpt is unknown, it is likely orders of magnitude greater than \longllama's three billion). Both \gpt and \llama outputs are affected by prompt variations, but \gpt's performance is more stable across prompts for all datasets. \gpt also has higher precision than \llama: \llama exhibits a non-zero false positive rate on the \dataset{Synthetic} dataset, which consists entirely of non-antisemitic event descriptions, whereas \gpt produces no false positives.

Third, our experiments on \dataset{AMCHA} reveal that model performance varies widely across fine-grained types and that there are heterogeneous effects of different prompting strategies. Specifically, including definitions of antisemitism and each category/type increases accuracy on rhetoric-oriented types (e.g. \bullying or use of \historical antisemitic tropes). Providing in-context examples increases accuracy on action-oriented types (e.g. physical \physical or \destruction of Jewish property). These findings suggest the need for context-sensitive model design that reflects the diverse ways in which harm is expressed, rather than relying on uniform prompts for identifying different kinds of hateful events.

Our results point to limitations of the current work and opportunities for future research. As our results demonstrate, there is substantial room for improvement in LLM performance on this task. Reproducibility and robustness remain limitations; there are numerous aspects of LLMs that could affect the models' performance on this task, potentially affecting any subsequent conclusions that may be drawn.
Even just between the two LLMs tested, the choice of model makes a large difference, and there are hundreds of other models that have not yet been evaluated on this task. We test several conceptually meaningful prompt augmentations, but there is still a vast space of seemingly minor prompt adjustments that could affect model performance, such as the specific format in which we request responses \citep{atreja2025s} and even minimal syntactic variations \citep{sclar2024quantifying}. For reproducibility, we opted to retain the same randomly-selected example per type for \oneshot prompting, but examples sampled more strategically could boost performance of this setup. Future work could systematically evaluate broader ranges of model sizes, architectures, and prompting strategies to understand how each of these variables influence performance. Fine-tuning LLMs (i.e., further training the LLM on labeled samples from our datasets and updating its internal weights) may also yield more reliable results, though there are high costs to such rigor in terms of money, computational resources, and environmental impact. Our experiments already cost \$350 to run \gpt with the OpenAI API (no web search or fine-tuning of weights, default chat completion feature), while \llama ran on 8 powerful A100 GPUs for about one day per dataset. Developing more efficient approaches will be crucial, not only for expanding the experimental scope, but also for scaling this pipeline for real-world deployment.

There are also several limitations in the datasets, taxonomies, and models that we analyze.
First, we believe that fine-grained taxonomies are an essential tool for antisemitic incident detection because definitions of antisemitism are highly contested and interpretations of incidents as antisemitic may be subjective. While no accepted definition labels all criticism of Israel as antisemitic, existing taxonomies rooted in the IHRA definition do include categories related to anti-Zionism (e.g. \textit{BDS activity} in the original AMCHA taxonomy). No matter what decision we make regarding these categories, disagreement and critique is inevitable. For the purpose of our experiments, we take a minimalist approach and exclude Israel-related categories in order to align with both the IHRA and JDA definitions of antisemitism \citep{ihra,jda}. However, we emphasize that our study does not advocate a normative stance, but rather presents a computational case study using a taxonomy and dataset suitable for LLM experimentation. Future research and practitioners should adapt categories as needed according to their own conceptual frameworks and analytic goals. 

While a strength of \dataset{AMCHA} is its curation by AMCHA Initiative team members, there is limited transparency regarding their internal data collection and labeling processes. For example, we do not have complete information on what (if any) submissions were rejected from inclusion into the corpus, or how what biases may have been introduced by their sampling strategy of manually tracking news, social media, and campus organizations. Some taxonomy decisions made by the AMCHA team, particularly treating \targeting and \expression as mutually exclusive and prioritizing the \targeting label at the expense of a balanced dataset, would benefit from greater conceptual clarity, as they have a major impact on LLM performance. There are potential biases in the other datasets as well: \dataset{ADL-HEAT} only includes incidents that have been reported to officials or submitted to the ADL. \dataset{Synthetic} relies on a small set of positive Jewish and Israel-related seed phrases and may not reflect real-world non-antisemitic reports; model performance could also be confounded by the fact that the texts are synthetic. Finally, \dataset{Campus-News} reflects specific newsroom editorial choices and may omit certain incidents, and the small size of annotated samples means that further validation is warranted. The taxonomies themselves are operationally useful for our study. Finally, our experiments only involve prompting LLMs without fine-tuning them on task-specific data. While we focus on in-context learning because it is most practical for the intended users of this antisemitism classification tool and because the size of the dataset is relatively small (especially broken down per category/type), further studies that compare fine-tuning and in-context learning on this event understanding task would yield additional insight.

Our findings may not generalize beyond the specific datasets and contexts examined. We focus on English-language, U.S.-based sources with an emphasis on antisemitic incidents on college campuses. Even within this context, our datasets necessarily contain a selection bias: they primarily include hateful events that have been reported and escalated to news media, civil society organizations, or law enforcement, but this constitutes a minority of the hateful events experienced by Jewish people \citep{adl2025_experience}. Future work could benefit from examining a more diverse range of data streams such as multiple social media platforms, potentially including not only textual descriptions but also other modalities such as audio, image, and video. Automated LLM-based pipelines for fine-grained hateful event detection should also be rigorously evaluated on expert-labeled data from other languages and countries. We also encourage future research to evaluate LLMs on this novel task on other hate ideologies besides antisemitism, which may each present their own unique challenges. 

The ability to conduct fine-grained hateful event detection opens several new promising research directions. One natural extension is to move beyond identifying hateful events themselves and further model how individuals and institutions respond to antisemitism in their communities, such as through counterspeech, policy actions, and public demonstrations of solidarity. Evaluating whether LLMs can reliably distinguish between descriptions of harm and responses to harm is itself an interesting and technically challenging problem \citep{gligoric-etal-2024-nlp}. More broadly, integrating hateful event detection with the identification of responses to hate and analyses of the online rhetoric that surrounds such events would enable a more holistic understanding of how harm is experienced, amplified, or mitigated across both online and offline contexts. Advancing this agenda will require closer collaboration between NLP researchers and practitioners to ensure that taxonomies, model assumptions, and outputs are aligned with real-world monitoring and intervention needs. 

Overall, our work shows that current LLMs are not yet sufficient as standalone systems, but can meaningfully augment human efforts to monitor and interpret hateful events at scale. Looking beyond antisemitism, this work points toward a future in which computational tools, used carefully and in partnership with domain expertise, help document social harm and support education, policy, and prevention efforts.

\appendix
\section{LLM Prompts}\label{subsec:prompts}
Example user prompts for each setting with article texts can be found in the \texttt{experiments/event\_discovery/log.txt} file in our GitHub repository, which will be released upon paper publication. Our system prompt and exact user prompt formulation function can be found in \texttt{experiments/event\_discovery/utils.py} within the same repository. In this appendix, we provide example prompt templates for each of our five settings. 
For \unforcednodef, our prompt template is as follows:
\begin{tcolorbox}[enhanced,breakable,width=\linewidth]
\begin{footnotesize}
\begin{Verbatim}[breaklines, breakanywhere,breaksymbol={}]
The following text, enclosed in single quotes, describes an event that happened on 2024-11-07 00:29:00 at Stanford University.

'[article text goes here]'

Does this quoted text describe an antisemitic event? If so, what category of antisemitism does the event fall under, and what type of expression does it fall under? Give an answer in the following format:

{'describes_antisemitic_event': bool,
'category': char,
'type_of_expression': int list}

For 'category', put the letter corresponding to the strongest option below for the quoted text:
A: 'Does not describe an antisemitic incident'
B: 'antisemitic expression'
C: 'targeting Jewish students and staff' 

For 'type_of_expression', list all numbers corresponding to the the following options that apply to the quoted text:
0: 'Does not describe an antisemitic incident'
1: 'physical assault' 
2: 'discrimination' 
3: 'destruction of Jewish property' 
4: 'genocidal expression'
5: 'suppression of speech/movement/assembly' 
6: 'bullying' 
7: 'denigration' 
8: 'historical' 
9: 'condoning terrorism'
\end{Verbatim}
\end{footnotesize}
\end{tcolorbox}
Our \forcednodef prompt template is as follows:
\begin{tcolorbox}[enhanced,breakable,width=\linewidth]
\begin{footnotesize}
\begin{Verbatim}[breaklines, breakanywhere,breaksymbol={}]
The following text, enclosed in single quotes, describes an event that happened on 2024-11-07 00:29:00 at Stanford University.

'[article text goes here]'

What category of antisemitism does the event fall under, and what type of expression does it fall under? Give an answer in the following format:

{'describes_antisemitic_event': bool,
'category': char,
'type_of_expression': int list}

For 'category', put the letter corresponding to the strongest option below for the quoted text:
A: 'Other type of expression'
B: 'antisemitic expression'
C: 'targeting Jewish students and staff' 

For 'type_of_expression', list all numbers corresponding to the the following options that apply to the quoted text:
0: 'Other type of expression'
1: 'physical assault' 
2: 'discrimination' 
3: 'destruction of Jewish property' 
4: 'genocidal expression'
5: 'suppression of speech/movement/assembly' 
6: 'bullying' 
7: 'denigration' 
8: 'historical' 
9: 'condoning terrorism'
If the best answer is 0, write a type of expression that you think would be the best fit in the 'other_type_of_expression' field. Otherwise, leave 'other_type_of_expression' as an empty string.
\end{Verbatim}
\end{footnotesize}
\end{tcolorbox}
For \unforceddef, our prompt template is as follows:
\begin{tcolorbox}[enhanced,breakable,width=\linewidth]
\begin{footnotesize}
\begin{Verbatim}[breaklines, breakanywhere,breaksymbol={}]
Antisemitism is defined as hostility to, prejudice towards, or discrimination against Jews. The following text, enclosed in single quotes, describes an event that happened on 2024-11-07 00:29:00 at Stanford University.

'[article text goes here]'

Does this quoted text describe an antisemitic event? If so, what category of antisemitism does the event fall under, and what type of expression does it fall under? Give an answer in the following format:

{'describes_antisemitic_event': bool,
'category': char,
'type_of_expression': int list}

For 'category', put the letter corresponding to the strongest option below for the quoted text:
A: 'Does not describe an antisemitic incident'
B: 'antisemitic expression' - Language, imagery or behavior deemed antisemitic by the U.S. State Department definition of antisemitism, or wholly consistent with that definition
C: 'targeting Jewish students and staff' - Incidents that directly target Jewish students on campus or other Jewish members of the campus community for harmful or hateful action based on their Jewishness or perceived support for Israel

For 'type_of_expression', list all numbers corresponding to the the following options that apply to the quoted text:
0: 'Does not describe an antisemitic incident'
1: 'physical assault' - Physically attacking Jewish students or staff because of their Jewishness or perceived association with Israel
2: 'discrimination' - Unfair treatment or exclusion of Jewish students or staff because of their Jewishness or perceived association with Israel
3: 'destruction of Jewish property' - Inflicting damage or destroying property owned by Jews or related to Jews
4: 'genocidal expression' - Using imagery (e.g. swastika) or language that expresses a desire or will to kill Jews or exterminate the Jewish people
5: 'suppression of speech/movement/assembly' - Preventing or impeding the expression of Jewish students, such as by removing or defacing Jewish students' flyers, attempting to disrupt or shut down speakers at Jewish or pro-Israel events, or blocking access to Jewish or pro-Israel student events
6: 'bullying' - Tormenting Jewish students or staff because of their Jewishness or perceived association with Israel
7: 'denigration' - Unfairly ostracizing, vilifying or defaming Jewish students or staff because of their Jewishness or perceived association with Israel
8: 'historical' - Using symbols, images and tropes associated with historical antisemitism, including by making "mendacious, dehumanizing, demonizing, or stereotypical allegations about Jews as such, or the power of Jews as a 
collective-especially but not exclusively, the myth about a world Jewish conspiracy or of Jews controlling the media, economy, governments, or other societal institutions"
9: 'condoning terrorism' - Calling for, aiding or justifying the killing or harming of Jews
\end{Verbatim}
\end{footnotesize}
\end{tcolorbox}
Our \forceddef prompt template is as follows:
\begin{tcolorbox}[enhanced,breakable,width=\linewidth]
\begin{footnotesize}
\begin{Verbatim}[breaklines, breakanywhere,breaksymbol={}]
Antisemitism is defined as hostility to, prejudice towards, or discrimination against Jews. The following text, enclosed in single quotes, describes an antisemitic event that happened on 2024-11-07 00:29:00 at Stanford University.

'[article text goes here]'

What category of antisemitism does the event fall under, and what type of expression does it fall under? Give an answer in the following format:

{'category': char,
'type_of_expression': int list}

For 'category', put the letter corresponding to the strongest option below for the quoted text:
A: 'Other type of expression'
B: 'antisemitic expression' - Language, imagery or behavior deemed antisemitic by the U.S. State Department definition of antisemitism, or wholly consistent with that definition
C: 'targeting Jewish students and staff' - Incidents that directly target Jewish students on campus or other Jewish members of the campus community for harmful or hateful action based on their Jewishness or perceived support for Israel

For 'type_of_expression', list all numbers corresponding to the the following options that apply to the quoted text:
0: 'Other type of expression'
1: 'physical assault' - Physically attacking Jewish students or staff because of their Jewishness or perceived association with Israel
2: 'discrimination' - Unfair treatment or exclusion of Jewish students or staff because of their Jewishness or perceived association with Israel
3: 'destruction of Jewish property' - Inflicting damage or destroying property owned by Jews or related to Jews
4: 'genocidal expression' - Using imagery (e.g. swastika) or language that expresses a desire or will to kill Jews or exterminate the Jewish people
5: 'suppression of speech/movement/assembly' - Preventing or impeding the expression of Jewish students, such as by removing or defacing Jewish students' flyers, attempting to disrupt or shut down speakers at Jewish or pro-Israel events, or blocking access to Jewish or pro-Israel student events
6: 'bullying' - Tormenting Jewish students or staff because of their Jewishness or perceived association with Israel
7: 'denigration' - Unfairly ostracizing, vilifying or defaming Jewish students or staff because of their Jewishness or perceived association with Israel
8: 'historical' - Using symbols, images and tropes associated with historical antisemitism, including by making "mendacious, dehumanizing, demonizing, or stereotypical allegations about Jews as such, or the power of Jews as a 
collective-especially but not exclusively, the myth about a world Jewish conspiracy or of Jews controlling the media, economy, governments, or other societal institutions"
9: 'condoning terrorism' - Calling for, aiding or justifying the killing or harming of Jews
If the best answer is 0, write a type of expression that you think would be the best fit in the 'other_type_of_expression' field. Otherwise, leave 'other_type_of_expression' as an empty string.
\end{Verbatim}
\end{footnotesize}
\end{tcolorbox}
Finally, our \oneshot prompt template is as follows:
\begin{tcolorbox}[enhanced,breakable,width=\linewidth]
\begin{footnotesize}
\begin{Verbatim}[breaklines, breakanywhere,breaksymbol={}]
Here are some examples of texts that describe antisemitic events, along with fine-grained labels of what types of antisemitism they describe:
Text: [text goes here]
Category: [ground truth category goes here]
Type of expression: [ground truth types go here]
The following text, enclosed in single quotes, describes an antisemitic event that happened on 2024-11-07 00:29:00 at Stanford University.

'[article text goes here]'

What category of antisemitism does the event fall under, and what type of expression does it fall under? Give an answer in the following format:

{'category': char,
'type_of_expression': int list}

For 'category', put the letter corresponding to the strongest option below for the quoted text:
A: 'Other type of expression'
B: 'antisemitic expression'
C: 'targeting Jewish students and staff'

For 'type_of_expression', list all numbers corresponding to the the following options that apply to the quoted text:
0: 'Other type of expression'
1: 'physical assault'
2: 'discrimination'
3: 'destruction of Jewish property'
4: 'genocidal expression'
5: 'suppression of speech/movement/assembly'
6: 'bullying'
7: 'denigration'
8: 'historical'
9: 'condoning terrorism'
If the best answer is 0, write a type of expression that you think would be the best fit in the 'other_type_of_expression' field. Otherwise, leave 'other_type_of_expression' as an empty string.
\end{Verbatim}
\end{footnotesize}
\end{tcolorbox}

An example article text, stripped of HTML, is as follows:
\begin{tcolorbox}
[enhanced,breakable,width=\linewidth]
\begin{footnotesize}
\begin{Verbatim}[breaklines, breakanywhere,breaksymbol={}]
"Heil Hitler" was found written on a poster-board in Harwood, a Pomona dorm hall.
\end{Verbatim}
\end{footnotesize}
\end{tcolorbox}

\texttt{gpt-4o} then returns the following response under the \unforcednodef variation:
\begin{tcolorbox}
[enhanced,breakable,width=\linewidth]
\begin{footnotesize}
\begin{Verbatim}[breaklines, breakanywhere,breaksymbol={}]
```json
{
  "describes_antisemitic_event": true,
  "category": "B",
  "type_of_expression": [4, 7, 9]
}
```
\end{Verbatim}
\end{footnotesize}
\end{tcolorbox}
Following the letter and number mappings in the question, this means that the model classifies this incident as ``antisemitic expression,'' falling under the types of ``genocidal expression,'' ``denigration,'' and ``condoning terrorism.''

It returns the following for \forcednodef:

\begin{tcolorbox}
[enhanced,breakable,width=\linewidth]
\begin{footnotesize}
\begin{Verbatim}[breaklines, breakanywhere,breaksymbol={}]
```json
{
  "category": "B",
  "type_of_expression": [4, 7, 9],
  "other_category": "",
  "other_type_of_expression": ""
}
```
\end{Verbatim}
\end{footnotesize}
\end{tcolorbox}

For \unforceddef, the response is as follows: 

\begin{tcolorbox}
[enhanced,breakable,width=\linewidth]
\begin{footnotesize}
\begin{Verbatim}[breaklines, breakanywhere,breaksymbol={}]
```json
{
  "describes_antisemitic_event": true,
  "category": "B",
  "type_of_expression": [4, 9]
}
```
\end{Verbatim}
\end{footnotesize}
\end{tcolorbox}

For \forceddef, the response is:
\begin{tcolorbox}
[enhanced,breakable,width=\linewidth]
\begin{footnotesize}
\begin{Verbatim}[breaklines, breakanywhere,breaksymbol={}]
```json
{
  "category": "B",
  "type_of_expression": [4, 9],
  "other_category": "",
  "other_type_of_expression": ""
}
```
\end{Verbatim}
\end{footnotesize}
\end{tcolorbox}
Finally, for \oneshot, the model returns: 
\begin{tcolorbox}
[enhanced,breakable,width=\linewidth]
\begin{footnotesize}
\begin{Verbatim}[breaklines, breakanywhere,breaksymbol={}]
```json
{
  "category": "B",
  "type_of_expression": [4, 7, 9],
  "other_category": "",
  "other_type_of_expression": ""
}
```
\end{Verbatim}
\end{footnotesize}
\end{tcolorbox}

\end{document}